\documentclass[sigconf,nonacm]{acmart}

\renewcommand\footnotetextcopyrightpermission[1]{}
\setcopyright{none}
\usepackage{amsmath}
\usepackage{booktabs}
\usepackage{tabularx}
\usepackage{array}
\usepackage{makecell}
\usepackage{cuted}
\usepackage{graphicx}
\usepackage{microtype}

\newcolumntype{Y}{>{\centering\arraybackslash}X}
\newcolumntype{L}{>{\raggedright\arraybackslash}X}
\newcolumntype{A}{>{\raggedright\arraybackslash\hsize=1.8\hsize}X}
\newcolumntype{B}{>{\centering\arraybackslash\hsize=.8\hsize}X}
\newcolumntype{C}{>{\raggedright\arraybackslash\hsize=1.65\hsize}X}
\newcolumntype{D}{>{\centering\arraybackslash\hsize=.5\hsize}X}
\newcolumntype{E}{>{\centering\arraybackslash\hsize=.75\hsize}X}
\newcolumntype{F}{>{\raggedright\arraybackslash\hsize=1.1\hsize}X}

\acmConference[Converted Manuscript]{Converted Manuscript}{2023}{Beijing, China}
\acmYear{2023}
\acmISBN{}
\acmDOI{}

\title{Topic Model Based on Co-occurrence Word Networks for Unbalanced Short Text Datasets}

\author{Chengjie Ma}
\email{macj163@163.com}
\affiliation{%
  \institution{Beijing Key Laboratory of Intelligent Telecommunication Software and Multimedia, Beijing University of Posts and Telecommunications}
  \city{Beijing}
  \country{China}
}

\author{Junping Du}
\authornote{Corresponding author.}
\email{junpingdu@126.com}
\affiliation{%
  \institution{Beijing Key Laboratory of Intelligent Telecommunication Software and Multimedia, Beijing University of Posts and Telecommunications}
  \city{Beijing}
  \country{China}
}

\author{Meiyu Liang}
\email{meiyu1210@bupt.edu.cn}
\affiliation{%
  \institution{Beijing Key Laboratory of Intelligent Telecommunication Software and Multimedia, Beijing University of Posts and Telecommunications}
  \city{Beijing}
  \country{China}
}

\author{Zeli Guan}
\email{guanzeli@bupt.edu.cn}
\affiliation{%
  \institution{Beijing Key Laboratory of Intelligent Telecommunication Software and Multimedia, Beijing University of Posts and Telecommunications}
  \city{Beijing}
  \country{China}
}

\begin{document}

\begin{abstract}
We propose a straightforward solution for detecting scarce topics in unbalanced short-text datasets. Our approach, named CWUTM (Topic model based on co-occurrence word networks for unbalanced short text datasets), addresses the challenge of sparse and unbalanced short text topics by mitigating the effects of incidental word co-occurrence. This allows our model to prioritize the identification of scarce topics (low-frequency topics). Unlike previous methods, CWUTM leverages co-occurrence word networks to capture the topic distribution of each word, and enhances the sensitivity in identifying scarce topics by redefining the calculation of node activity and normalizing the representation of both scarce and abundant topics to some extent. Moreover, CWUTM adopts Gibbs sampling, similar to LDA, making it easily adaptable to various application scenarios. Extensive experimental validation on unbalanced short-text datasets demonstrates the superiority of CWUTM compared to baseline approaches in discovering scarce topics. According to the experimental results, the proposed model is effective in early and accurate detection of emerging topics or unexpected events on social platforms.
\end{abstract}


\thanks{This work was supported by the Program of the National Natural Science Foundation of China (62192784, U22B2038, 62172056) and by Young Elite Scientists Sponsorship Program by CAST (2022QNRC001).}

\maketitle

\section{Introduction}

Topic models serve as statistical tools used to uncover concealed semantic structures within document collections \cite{blei2003latent}. These models, along with their extensions, have found applications in diverse fields including marketing, sociology, political science, among others \cite{boydgraber2017applications}. In applied intelligent decision scenarios, interpretable machine learning also emphasizes that learned semantic features should remain transparent and actionable \cite{li2019interpretable}.

Most topic models are advancements built upon the latent Dirichlet allocation (LDA) technique \cite{blei2003latent}. LDA, being the canonical form of existing topic models, employs a hierarchical parametric Bayesian approach to uncover topics within extensive corpora. It represents documents as a mixture of topics, with each topic being a probability distribution of words from the corpus vocabulary. Through statistical inference, LDA learns the probability distribution of words associated with each topic and the topic distribution for each document. However, LDA-like models, which leverage document-level word co-occurrence information \cite{wang2006topics}, tend to consolidate semantically related words into a single topic. This characteristic makes them highly sensitive to the length and quantity of documents attributed to each topic. Consequently, when working with short texts that contain a limited number of words, these models fail to capture the relationships among words.

With the rapid evolution of the World Wide Web and the emergence of various web applications, short texts have become predominant content on the Internet \cite{kou2018hashtag,wei2019boosting,li2022fuel}. For instance, Twitter's user base generates a vast number of tweets daily, carrying valuable signals that reflect the real world. Accurately extracting topics from these short texts is crucial for tasks like topic detection \cite{wang2007mining}, query suggestions, user interest monitoring \cite{weng2010twitterrank}, document classification, comment summarization \cite{ma2012topic}, and text clustering. Heterogeneous graph attention networks further show that typed semantic relations can strengthen semi-supervised short text classification when labels are limited \cite{hu2019hgat}. However, short texts suffer from sparsity and lack of word co-occurrence details, making traditional topic models struggle when applied to them. This has sparked interest in the machine learning research community, focusing on short text topic modeling to overcome the sparsity issue.

Indeed, numerous attempts have been made to address the limitations of LDA when modeling short texts \cite{yuan2020financial,xiao2022lecf,li2018resilient,huang2021hgamn}. One approach involves aggregating related short texts into longer pseudo-documents prior to training topic models \cite{weng2010twitterrank}. Another strategy utilizes models trained on external data sources such as Wikipedia to enhance thematic understanding in short texts \cite{xuan2008learning}. Furthermore, several LDA operations have been introduced with the aim of achieving optimal performance in short texts \cite{chen2013emerging,chua2013automatic,zhao2011comparing,meng2013tracking,li2017tobit}. In 2013, a specialized hybrid model called the BITERM topic model emerged as an alternative to LDA for short texts \cite{yan2013biterm}. Although the BITERM model exhibits proficiency in handling short texts, it does not overcome the shortcomings of LDA-like methods, thus limiting its flexibility. Another notable example is the double sparse topic model \cite{lin2014dual}, which modifies LDA to capture the primary topic of each short document and the key terms associated with each topic. Additionally, in recent years, there has been a rise in methods combining neural networks to tackle short text topic modeling \cite{feng2020context}. Retrieval-oriented pre-training, such as masked auto-encoder learning for dense retrieval, also indicates that language representations can be specialized for sparse matching scenarios \cite{xiao2022retromae}.

In the field of short text topic modeling, there are significant practical applications for identifying rare topics in unbalanced datasets \cite{shao2021memory,yu2023fedpcf,li2022crossmedia,li2022distributed}. For instance, the timely discovery of crisis events on social platforms can greatly minimize losses, while the early detection of scarce topics can enhance the prediction of future events. Federated supervised cross-modal retrieval is another related setting for aligning heterogeneous scientific and technological information without centralizing all resources \cite{li2024fedCrossModal}. Adjacent recommendation research has also explored lightweight sequence representation through filter-enhanced MLPs, showing the value of compact feature filtering in sparse behavioral data \cite{zhou2022filterMLP}. Dataset distillation for sequential recommendation is less directly tied to topic modeling, but it similarly studies how compact data can preserve learning signals in sparse sequential domains \cite{zhang2025td3}. This model specifically focuses on scarce topics in unbalanced datasets within the context of short text modeling.

This paper primarily aims to enhance traditional word co-\allowbreak occurrence by leveraging network-based approaches. It improves the calculation of word co-\allowbreak occurrence network activity and implements appropriate pruning techniques to mitigate the impact of random word pairings within documents. This enhancement enables the model to effectively identify scarce topics in unbalanced texts while reducing the overall model complexity. Subsequently, the word co-occurrence network is transformed back into a pseudo-document set, and topics are extracted from the co-occurrence network using Gibbs sampling, similar to the approach used in LDA for solving text topics.

\section{Related Work}

Probabilistic latent semantic indexing (PLSA) \cite{hofmann1999plsi} and latent Dirichlet allocation (LDA) have been extensively employed in text corpus research. LDA, in particular, is considered a more comprehensive generative model as it extends PLSA by incorporating Dirichlet priors on topic distributions. Over the past two decades, various complex variants of LDA and PLSA have emerged due to their extensibility, including dynamic topic models, social topic models, author-topic models, and author-community models \cite{authorTopicCommunity2015}. Scientific publication representation learning further shows that semantic-similarity attention and hypergraph convolution can model higher-order relations among scholarly objects \cite{li2026semanticHypergraph}.

In recent years, the proliferation of the Internet has resulted in a significant increase in text data, garnering increased attention for research in this field \cite{wu2020short,li2017variance,meng2015robust,cao2013review}. Federated graph neural networks have studied cross-graph node classification under separated graph data, which is relevant to distributed representation learning \cite{guan2021federated}. Early studies primarily focused on leveraging auxiliary information to enhance short data density \cite{meng2015formation,li2023latenttopic}. Multi-view scholar clustering with dynamic interest tracking further shows that scientific resources can be multi-view and temporally changing \cite{li2023scholar}. For instance, Wang et al. \cite{wang2007mining} trained topic models on aggregated tweets that shared common words, which yielded better results compared to models trained directly on the original tweets. Wang \cite{wang2006topics} proposed a method for measuring similarity in short texts based on search fragments. Chen et al. \cite{chen2013emerging} introduced a specialized form of hybrid model to enhance topic modeling of short texts. Zhao et al. \cite{zhao2011comparing} incorporated sparse constraints on document-topic distribution and topic-term distribution to model topics in short texts. Feng et al. \cite{feng2020context} combined neural network techniques with other methods to tackle short text topic modeling, while Van et al. \cite{van2022graph} utilized graph convolutional neural networks to address short text problems. T2-GNN and reciprocally contrastive heterogeneous graph learning further indicate that incomplete graph features and multiple graph views can be exploited rather than discarded \cite{huo2023t2gnn,jin2022heterogeneousGraphContrastive}. Federated self-adaptive learning for information networks provides another privacy-aware representation setting when data sources cannot be directly centralized \cite{li2026fedsin}.

To handle unbalanced texts in topic modeling, prior knowledge has been widely utilized to alleviate the skewed distribution of documents across different topics. Andrzejewski et al. \cite{andrzejewski2009domain} suggested incorporating ``must link'' and ``cannot link'' constraints into the topic model. Allahyari and Kochut \cite{allahyari2016coherent} employed general vocabulary knowledge to aid in the discovery of coherent themes. Broader graph learning studies also provide useful perspectives: modularity-based deep learning mines community structures \cite{yang2016modularity}, and self-supervised graph co-training learns from complementary graph views \cite{xia2021graphCoTraining}. However, there is still a lack of a definitive solution for topic modeling of unbalanced texts.

\section{Topic Model Based on Co-occurrence Word Networks for Unbalanced Short Text Datasets}

In this section, we present a topic model specifically designed for unbalanced short text datasets, utilizing co-occurrence word networks. This model addresses the unbalance issue by normalizing both scarce and abundant topics to some extent. It leverages the construction of co-occurrence word networks to effectively capture context information. Furthermore, the model employs techniques inspired by LDA for topic prediction. By combining these approaches, we aim to enhance the performance of topic modeling on unbalanced short text datasets.

\subsection{Co-occurrence Word Network}

\begin{figure}
  \centering
  \includegraphics[width=\columnwidth]{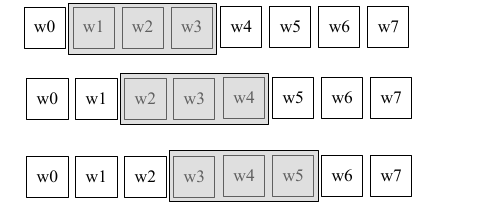}
  \caption{Sliding window process.}
  \Description{Three sliding windows over a sequence of words from w0 to w7.}
  \label{fig:sliding-window}
\end{figure}

In a word co-occurrence network (referred to as a word network hereafter, unless stated otherwise), nodes represent words present in a given corpus, while the edges connecting the nodes indicate that the corresponding words have co-occurred at least once within the same context. The context can refer to either a document or a fixed-size sliding window. To maintain the size of the word network and capture the local context of each word effectively, this study adopts sliding windows of a predetermined size as the context of analysis. Research suggests that sliding windows of size 10 or larger can capture topic similarity between words comprehensively \cite{zuo2016wntm}. However, larger window sizes result in higher computational complexity, as depicted in Figure~\ref{fig:sliding-window}. Hence, in this paper, we have opted to set the sliding window size as 10 for normal text and short text.

To convert a given set of documents into a word network, several steps are followed. Firstly, low-frequency words and stop words are filtered out from the documents. Then, a sliding window is employed to scan each document. This sliding window moves word by word, and whenever two distinct words appear within the same window, they are considered to co-occur. The weight of the corresponding edge between the two words is determined by the cumulative count of their co-occurrences. It is important to note that a word pair may be counted multiple times, as depicted in the word-pair weighting pattern illustrated in Figure~\ref{fig:sliding-window}. Words that appear in adjacent positions within the window are counted more frequently compared to words that appear further apart. For instance, in the given example, the pair of words W2 and W3 is counted twice, whereas the pair W0 and W7 is counted only once. This approach facilitates the models to associate adjacent words within the same topic, thereby enhancing the learning of topic coherence. Cohesive words that are semantically strongly related are often found in close proximity to each other, making the adjacency-based grouping beneficial for topic modeling.

In the proposed topic model, a topic can be viewed as a collection of words that frequently co-occur within the same document or time window. These co-occurring words form potential phrases that exhibit similarity in the word network or community. The reason for their similarity is that when words appear frequently within the same sliding window or document, they are more likely to be closely related in the semantic space. Therefore, in the LDA-based model, we can utilize these potential phrases extracted from the word network as topics. Simultaneously, the model learns the specific representation of topics from the word co-occurrence network, which offers a theoretical basis for topic coherence, as discussed in \cite{authorTopicCommunity2015}.

The conventional approach for calculating the weight of word network nodes involves using the frequency of co-occurrence between two nodes as their weight. However, in the case of abundant topics in an unbalanced text dataset, where topic-related texts occur frequently, the weights of two words can be high even if they do not always co-occur. On the other hand, for words from scarce topics, their weights remain low despite consistent co-occurrence. This traditional method fails to capture the association of scarce topics effectively. To address this issue and account for incidental co-occurrence, we propose a redefined approach to calculate the weight between nodes. This ensures that our model can accurately identify scarce topic texts in unbalanced text datasets. The activity degree between nodes $\langle w_x, w_y \rangle$ is calculated as follows:

\begin{equation}
\operatorname{degree}(w_x,w_y)=\max\left[\log\left(\frac{p(w_x,w_y)}{p(w_x)p(w_y)}\right),0\right].
\label{eq:degree}
\end{equation}

\begin{figure}
  \centering
  \includegraphics[width=\columnwidth]{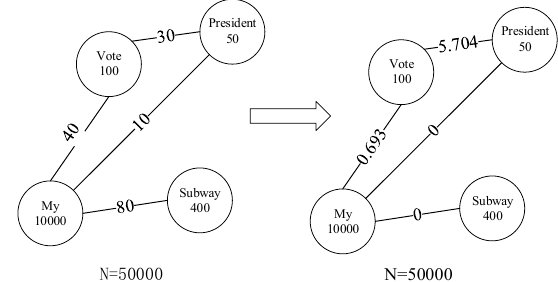}
  \caption{The left figure represents the weights in the basic co-occurrence network, which reflects the raw co-occurrence frequencies between words. The right figure illustrates the calculation method employed in this model to determine the co-occurrence network.}
  \Description{Two small word networks comparing raw co-occurrence frequencies and the proposed activity-weighted network.}
  \label{fig:weights}
\end{figure}

By setting the activity level in this manner, the influence of chance occurrences of two words appearing together can be mitigated to some extent. Additionally, the impact of word frequency on topic recognition can be attenuated, which is particularly useful for unbalanced text datasets. When the activity level is zero, $P(w_x,w_y)=p(w_x)P(w_y)$, indicating that the occurrence of the two words is independent of each other and the link between the two nodes is cancelled. Positive values indicate that the words co-occur more frequently than expected by chance, suggesting that the two words are more likely to belong to the same topic. Negative values indicate that when one word appears, the other word is less likely to appear, implying that the two words should not be assigned to the same topic. This approach also facilitates the pruning of the original word co-occurrence network, thereby reducing the model's complexity. The results of this calculation method are illustrated in Figure~\ref{fig:weights}.

\subsection{Discover Topics from the Co-occurrence Network}

To uncover topics from co-occurrence networks, we adopt the method used in the Word Network Topic Model (WNTM) \cite{allahyari2016coherent,zuo2016wntm} and employ standard Gibbs sampling. The first step is to represent the word co-occurrence network as a pseudo-document set. We assume that the list of adjacent words for each word is generated semantically according to a specific probability model, which allows us to learn the statistical relationships between words, potential word groups, and the lists of adjacent words for each word.

We make the initial assumption that there exists a fixed set of potential phrases in the lexical network, and each potential phrase $Z$ is associated with a multinomial distribution on the vocabulary $\Phi_z$. This multinomial distribution is derived from a Dirichlet prior $\operatorname{Dir}(\beta)$. The generation process for the entire pseudo-document set, which is transformed from the word network, can be described as follows:

\begin{itemize}
  \item For each potential word group $Z$, the polynomial distribution of $Z$ in the word group $\Phi_z \sim \operatorname{Dir}(\beta)$ is obtained.
  \item The Dirichlet distribution $\theta_i \sim \operatorname{Dir}(\alpha)$ of the potential word group of the adjacent word list $L_i$ of the word $w_i$ is obtained.
  \item For each word $w_j \in L_i$: select a potential phrase $z_j \sim \theta_i$. Then select an adjacency $w_j \sim \Phi_{z_j}$.
\end{itemize}

In our model, the $\theta$ distribution represents the probability of the latent phrase appearing in the adjacent word list of each word, while the $\Phi$ distribution represents the probability of a word belonging to each potential phrase. Given the observed corpus, the model initially converts it into a co-occurrence word network. From this network, a pseudo-document set is generated. Subsequently, the model utilizes the same Gibbs sampling implementation as traditional latent Dirichlet allocation to infer the values of potential variables in $\Phi$ and $\theta$. This inference process allows us to estimate the probabilities associated with the latent phrases and word assignments within the model.

\begin{figure*}
  \centering
  \includegraphics[width=.88\textwidth]{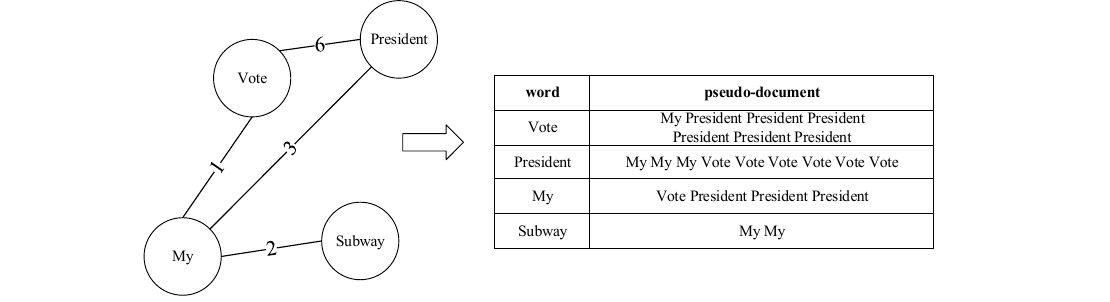}
  \caption{The process of generating virtual text from a co-occurrence network.}
  \Description{A small co-occurrence network is transformed into a pseudo-document table.}
  \label{fig:virtual-text}
\end{figure*}

When inferring topics in short texts, we can utilize the topic proportions of words to their adjacent word lists $\theta_i$ as the topic proportions in $w_i$. By obtaining the topic proportions for all words, we can then determine the topic proportions for each document. This can be expressed as follows:

\begin{equation}
P(z\mid d)=\sum_{w_i} P(z\mid w_i)P(w_i\mid d).
\label{eq:doc-topic}
\end{equation}

$P(z\mid w_i)=\theta_{i,z}$, the empirical distribution of the document word, is estimated as

\begin{equation}
P(w_i\mid d)=\frac{n_d(w_i)}{\operatorname{Len}(d)}.
\label{eq:word-doc}
\end{equation}

Here, $n_d(w_i)$ is the word frequency of $w_i$ in document $d$, and $\operatorname{Len}(d)$ is the length of $d$. The methods described above provide a straightforward approach to infer the topic of a given passage.

\section{Experiments}

\subsection{Datasets}

In this section, we introduce the datasets and evaluation metrics used in our experiments.

\textbf{Short text datasets.} We selected three datasets \cite{qiang2020survey}, namely SearchSnippets, Tweet, and GoogleNews, to validate the effectiveness of our model. The key information of each dataset, including the number of topics (K), number of documents (N), average and maximum document length (Len), and vocabulary size (V), is summarized in Table~\ref{tab:datasets}.

\begin{table}
  \caption{Basic information about datasets.}
  \label{tab:datasets}
  \centering
  \begin{tabularx}{\columnwidth}{@{}A BBBB@{}}
    \toprule
    Dataset & K & N & Len & V \\
    \midrule
    Search Snippets & 8 & 12,295 & 14.4/37 & 5,547 \\
    Tweet & 89 & 2,472 & 8.55/20 & 5,096 \\
    Google News & 152 & 11,108 & 6.23/14 & 8,110 \\
    \bottomrule
  \end{tabularx}
\end{table}

\textbf{Unbalanced short text datasets.} The three datasets were processed and partitioned into two subsets: scarce topics (low-\allowbreak frequency topics) and abundant topics (high-\allowbreak frequency topics), based on the number of texts within each topic. In the table below, K and N are the same as above, and Table~\ref{tab:unbalanced-datasets} displays the number of texts on this topic within the respective subset.

\begin{table}
  \caption{Basic information about unbalanced datasets.}
  \label{tab:unbalanced-datasets}
  \centering
  \begin{tabularx}{\columnwidth}{@{}C D E F@{}}
    \toprule
    Dataset & K & N & Number of texts on this topic \\
    \midrule
    Scarce/Abundant Google News & 99/53 & 2978/8130 & $\leq$80 / 153.40 (average) \\
    Scarce/Abundant Search Snippets & 2/6 & 1544/10751 & $<$1200 / 1703.71 (average) \\
    Scarce/Abundant Tweet & 49/40 & 335/2137 & $\leq$15 / 53.43 (average) \\
    \bottomrule
  \end{tabularx}
\end{table}

\subsection{Evaluation Metrics}

To evaluate the performance of the model in clustering the documents, the following measurement indicators are used: Purity and NMI (normalized mutual information) \cite{andrzejewski2009domain,huang2013dirichlet}.

\textbf{Normalized mutual information (NMI).} NMI is a statistical measure that evaluates the similarity between the cluster labels and the real labels, taking into account the distribution of documents across clusters and classes. It calculates the mutual information between the two sets of labels, normalized by the entropy of the labels. NMI ranges from 0 to 1, with higher values indicating better agreement between the cluster labels and the real labels.

\begin{equation}
NMI=\frac{\sum_{h,l} d_{hl}\log\left(\frac{D\cdot d_{hl}}{d_h c_l}\right)}
{\sqrt{\left(\sum_h d_h\log\left(\frac{d_h}{D}\right)\right)\left(\sum_l c_l\log\left(\frac{c_l}{D}\right)\right)}}.
\label{eq:nmi}
\end{equation}

Here, $D$ is the number of documents, $d_h$ is the number of documents in class $h$, $c_l$ is the number of documents in group $l$, and $d_{hl}$ is the number of documents in class $h$ and group $l$. The NMI value is 1 when the clustering solution exactly matches the user-flagged category assignment, and is close to 0 for random document partitioning.

\textbf{Clustering purity.} Purity measures the quality of clustering results by comparing the cluster labels assigned by the model with the real labels of the documents. It calculates the proportion of correctly assigned documents to the total number of documents. Higher purity values indicate better clustering performance. The calculation formula of purity is defined as follows:

\begin{equation}
P(\Omega,C)=\frac{1}{N}\sum_k \max_j |\omega_k \cap c_j|.
\label{eq:purity}
\end{equation}

Here, $N$ represents the total number of samples. The cluster set and the correct category set are
\[
\Omega=\{\omega_1,\omega_2,\ldots,\omega_K\}, \qquad
C=\{c_1,c_2,\ldots,c_j\}.
\]
$\omega_k$ represents all samples in the $k$th cluster after clustering, and $c_j$ represents the real samples in the $j$th category. Here, the value range of $P$ is $[0,1]$. The larger the value, the better the clustering effect is.

\subsection{Parameter Setting}

LDA and WNTM were used as comparison models to test the performance of this model. Because LDA is the basic model in the topic model and WNTM is the original model of this model, this model improves its recognition performance of scarcity categories in unbalanced data sets on the basis of WNTM model. The results of this experiment are averaged after 10 runs.

\textbf{LDA:} Through grid search, the hyperparameters of LDA are $\alpha=0.05$ and $\beta=0.01$ \cite{qiang2018sttm} to obtain the best performance.

\textbf{WNTM:} For this model, use the hyperparameters $\alpha=0.1$, $\beta=0.1$ from the original paper and set the window size to 10 words for best performance.

\textbf{CWUTM:} For this model, use $\alpha=0.1$, $\beta=0.1$ and set the window size to 10 words to get better performance.

\subsection{Models in General Datasets}

Through experiments, the results are shown in Table~\ref{tab:general}. Based on the experimental findings, it is evident that the CWUTM model exhibits lower performance compared to the benchmark model in normal datasets. Further analysis reveals that this discrepancy stems from the model's inclination towards scarce topics, thereby neglecting the abundant topics. Additionally, the calculation of the overall clustering evaluation index assigns greater weightage to the abundant topics, resulting in a decrease in the overall performance of the CWUTM model, despite its proficiency in handling scarce topics.

\begin{strip}
  \centering
  \captionof{table}{Model performance of each model in general datasets.}
  \label{tab:general}
  \begin{tabularx}{\textwidth}{@{}L L Y Y Y Y@{}}
    \toprule
    Model & Metric & Google News (\%) & Search Snippets (\%) & Tweet (\%) & Mean Value (\%) \\
    \midrule
    LDA \cite{blei2003latent} & Purity & 78.89 ($\pm$1.25) & 69.55 ($\pm$4.43) & 78.32 ($\pm$1.08) & \textbf{75.59} \\
     & NMI & 82.22 ($\pm$0.47) & 48.25 ($\pm$3.13) & 78.02 ($\pm$0.70) & 69.50 \\
    WNTM \cite{zuo2016wntm} & Purity & \textbf{83.20} ($\pm$0.97) & 59.33 ($\pm$4.01) & \textbf{82.40} ($\pm$0.88) & 74.98 \\
     & NMI & \textbf{89.63} ($\pm$0.28) & 43.72 ($\pm$3.58) & \textbf{82.72} ($\pm$0.67) & \textbf{72.02} \\
    CWUTM (Ours) & Purity & 81.31 ($\pm$1.28) & 62.43 ($\pm$2.95) & 74.79 ($\pm$1.28) & 72.84 \\
     & NMI & 86.17 ($\pm$0.77) & 46.30 ($\pm$3.11) & 75.86 ($\pm$0.88) & 69.44 \\
    \bottomrule
  \end{tabularx}
\end{strip}

However, it is crucial to highlight that the CWUTM model excels in the identification of scarce topics, as will be elaborated upon in subsequent sections. While the model may exhibit limitations in normal datasets, it demonstrates the advantage of effectively recognizing and clustering sparse topics in unbalanced text datasets.

\subsection{Models in Imbalanced Datasets}

Through experimentation on the unbalanced datasets, we obtained results that highlight the behavior of different models. Specifically, the scarce subset is designed to evaluate the performance of the model in identifying scarce topics, where the number of texts per topic is relatively low. On the other hand, the abundant subset serves as a comparison experiment. The experimental findings are summarized in Tables~\ref{tab:scarce} and~\ref{tab:abundant}.

\begin{strip}
  \centering
  \captionof{table}{Model performance of each model in scarce subsets.}
  \label{tab:scarce}
  \begin{tabularx}{\textwidth}{@{}L L Y Y Y Y@{}}
    \toprule
    Model & Scarce subsets & Google News (\%) & Search Snippets (\%) & Tweet (\%) & Mean Value (\%) \\
    \midrule
    LDA \cite{blei2003latent} & Purity & 78.25 ($\pm$1.76) & \textbf{92.34} ($\pm$1.58) & 75.94 ($\pm$2.11) & 82.18 \\
     & NMI & 88.64 ($\pm$1.03) & 39.80 ($\pm$4.81) & 85.94 ($\pm$0.81) & 71.46 \\
    WNTM \cite{zuo2016wntm} & Purity & 77.84 ($\pm$1.62) & 87.57 ($\pm$6.37) & 75.28 ($\pm$3.20) & 80.23 \\
     & NMI & 90.96 ($\pm$0.37) & 31.22 ($\pm$15.98) & \textbf{86.88} ($\pm$0.96) & 69.69 \\
    CWUTM (Ours) & Purity & \textbf{81.49} ($\pm$2.94) & 92.26 ($\pm$2.48) & \textbf{76.60} ($\pm$1.74) & \textbf{83.45} \\
     & NMI & \textbf{91.38} ($\pm$0.74) & \textbf{41.79} ($\pm$7.82) & 85.63 ($\pm$0.98) & \textbf{72.94} \\
    \bottomrule
  \end{tabularx}
  \captionof{table}{Model performance of each model in abundant subsets.}
  \label{tab:abundant}
  \begin{tabularx}{\textwidth}{@{}L L Y Y Y Y@{}}
    \toprule
    Model & Abundant subsets & Google News (\%) & Search Snippets (\%) & Tweet (\%) & Mean Value (\%) \\
    \midrule
    LDA \cite{blei2003latent} & Purity & 86.51 ($\pm$0.55) & \textbf{73.21} ($\pm$3.99) & 88.38 ($\pm$0.81) & \textbf{82.70} \\
     & NMI & 80.80 ($\pm$0.49) & \textbf{47.35} ($\pm$3.47) & 80.05 ($\pm$0.45) & 69.40 \\
    WNTM \cite{zuo2016wntm} & Purity & \textbf{90.09} ($\pm$0.96) & 61.46 ($\pm$3.14) & \textbf{89.42} ($\pm$1.44) & 80.33 \\
     & NMI & \textbf{89.92} ($\pm$0.25) & 42.16 ($\pm$2.63) & \textbf{83.38} ($\pm$1.23) & \textbf{71.82} \\
    CWUTM (Ours) & Purity & 87.40 ($\pm$1.17) & 63.95 ($\pm$1.37) & 82.55 ($\pm$1.23) & 77.96 \\
     & NMI & 84.28 ($\pm$1.45) & 43.93 ($\pm$1.55) & 75.69 ($\pm$0.62) & 68.03 \\
    \bottomrule
  \end{tabularx}
\end{strip}

The results demonstrate that the CWUTM model performs well in identifying scarce topics compared to the LDA and WNTM models. Specifically, it achieves higher purity accuracy and NMI scores on the subset of scarce topics in all three datasets. On average, the CWUTM model outperforms the LDA model by 1.27\% in purity accuracy and 1.48\% in NMI, while surpassing the WNTM model by 3.22\% in purity accuracy and 3.25\% in NMI.

However, there is a trade-off as the LDA and WNTM models outperform the CWUTM model when it comes to the abundant subsets of the datasets. This suggests that the CWUTM model places a greater emphasis on scarce topics, which affects its ability to cluster abundant topics. This imbalance in emphasis ultimately impacts its overall performance on the dataset. On the other hand, the LDA and WNTM models allocate more attention to abundant topics, resulting in higher overall performance.

It is worth noting that in the context of the web, where there are emerging or emergent events with limited relevant text, traditional topic models often struggle to capture information on scarce topics. In such cases, our CWUTM model may provide better performance and be more effective in identifying and analyzing scarce topics.

\section{Conclusion}

CWUTM leverages the co-occurrence word network to model the topic distribution of each word, thereby enhancing the semantic density of the data space. By refining the calculation method of node activities, CWUTM effectively normalizes the representation of both scarce and large topics, ensuring the sensitivity of the data space to identify rare topics. While the model may exhibit lower overall performance due to the potential loss of data related to large topics, it demonstrates excellent performance in capturing and recognizing rare topics.

The strengths of CWUTM lie in its ability to efficiently and accurately discover emerging topics or unexpected events on social platforms. The model's focus on scarce topics makes it particularly suitable for early detection and precise identification of emerging trends or unusual occurrences.

\bibliographystyle{gbt7714-numerical}
\bibliography{references}

\end{document}